\documentclass[conference]{IEEEtran}

\usepackage{algorithm}
\usepackage{algpseudocode}
\usepackage{amsmath}
\usepackage{amssymb}
\usepackage{color}
\usepackage{lettrine}

\begin{document}

\title{Chemical Reaction Optimization for the Set Covering Problem}

\author{James J.Q. Yu,
       \textit{Student Member, IEEE}\\
       Department of Electrical and\\
       Electronic Engineering\\
       The University of Hong Kong\\
       Email: jqyu@eee.hku.hk\\
\and Albert Y.S. Lam,
       \textit{Member, IEEE}\\
       Department of Computer Science\\
       Hong Kong Baptist University\\
       Email: albertlam@ieee.org\\
\and Victor O.K. Li,
       \textit{Fellow, IEEE}\\
       Department of Electrical and\\
       Electronic Engineering\\
       The University of Hong Kong\\
       Email: vli@eee.hku.hk\\
}

\maketitle
\pagestyle{empty}

\begin{abstract}
The set covering problem (SCP) is one of the representative combinatorial optimization problems, having many practical applications. This paper investigates the development of an algorithm to solve SCP by employing chemical reaction optimization (CRO), a general-purpose metaheuristic. It is tested on a wide range of benchmark instances of SCP. The simulation results indicate that this algorithm gives outstanding performance compared with other heuristics and metaheuristics in solving SCP.
\end{abstract}


\section{Introduction}

\lettrine[lines=2]{T}{he set} covering problem (SCP) is one of the representative combinatorial optimization problems. It has many real-world applications, e.g. bus, railway and airline crew scheduling, vehicle routing, facility location, and political districting \cite{BalasPadberg1976SetPartitioning:Survey}. More recent applications of SCP are on sensor lifetime maximization \cite{HuZhangYuChungLiShiLuo2010HybridGeneticAlgorithm} and phasor measurement unit placement \cite{XuWenLeungLi2013OptimalPMUPlacement}.

SCP is formally defined as follows. We have a set of $m$ elements $\mathbb{M}=\{1,\cdots,m\}$ and a collection of $n$ subsets $\mathbb{N} = \{S_j\subseteq\mathbb{M},1\leq j\leq n\}$, each of which is associated with a cost $S_j$, denoted as $c_j$. We say a collection of subsets $\mathbb{X}\subseteq\mathbb{N}$ is a cover of $\mathbb{M}$ if $\bigcup_{S_j\in\mathbb{X}}S_j=\mathbb{M}$ holds. $\mathbb{X}$ is a prime cover of $\mathbb{M}$ if there is no redundant subset in $\mathbb{X}$, i.e., $\mathbb{X}$ will not cover $\mathbb{M}$ if any subset is removed from $\mathbb{X}$. The goal of SCP is to find an $\mathbb{X}$ with the minimum cost.

SCP is usually formulated as a binary integer programming problem as follows:
\begin{equation}
\begin{aligned}
& \min && \sum^n_{j=1}c_jx_j \\
& \text{s.t.} && \sum^n_{j=1}a_{ij}x_j\geq1, \; i=1,2,\cdots,m, \\
& && x_j\in\{0,1\}, \; j=1,2,\cdots,n,
\end{aligned}
\end{equation}
where $a_{ij}=1$ if $i\in S_j$ and $a_{ij}=0$ otherwise. The decision variable $x_{j}$ is set to one if subset $S_j$ is selected in the cover $\mathbb{X}$.

It is also common to formulate SCP into matrix form. In this formulation, SCP is the problem of covering the rows of an $m\times n$ matrix by a subset of the columns at a minimal cost. We use $\mathbf{A}=\{a_{ij}, 1\leq i\leq m, 1\leq j\leq n\}$ to represent the matrix, and we say the $k$-th element is covered by the $l$-th subset if $a_{kl}=1$. We use $\mathsf{C}=\{c_j,1\leq j\leq n\}$ as the cost coefficient vector. Then SCP is defined as follows:
\begin{equation}\label{eqn:form2}
\begin{aligned}
& \min && \mathsf{C^\top X} \\
& \text{s.t.} && \mathbf{A}\mathsf{X}\geq\mathsf{b}, \\
\end{aligned}
\end{equation}
where $\mathsf{X}=\{x_j,1\leq j\leq n\}$ is the solution vector and $\mathsf{b}$ is the unit vector of length $m$.  The constraint ensures that each row is covered by at least one column. If the costs for all subsets are identical, then the SCP is named \textit{unicost} SCP. Otherwise, it is called \textit{weighted} or \textit{non-unicost} SCP.

SCP is known to be an NP-hard optimization problem \cite{GareyJohnson1979ComputersandIntractability:}, and metaheuristics have been shown to be effective in solving complex problems. Chemical reaction optimization (CRO) is one of the general-purpose metaheuristics which has shown its capability in solving similar NP-hard combinatorial problems, e.g. the quadratic assignment problem \cite{LamLi2010ChemicalReactionInspired}. CRO is inspired by chemical reactions, where reactant molecules collide with container walls or with each other. During the collision, the structure of these molecules may change and the energy hold by these molecules may be transferred to other molecules or transformed into other energy forms. The changes made by the collision follow a natural tendency that the potential energy of product molecules is smaller than that of reactant molecules macroscopically \cite{LamLi2010ChemicalReactionInspired}. CRO utilizes this tendency to perform optimization.

In this paper, we propose a heuristic-based CRO algorithm to solve SCP, named hCRO. We perform a series of simulations to test its performance. This paper is organized as follows. We perform a brief survey on previously proposed approaches to solve SCP, and some prior efforts on CRO in Section II. Section III presents the design and implementation of our proposed hCRO algorithm. The performance of our proposed algorithm is illustrated with the help of a series of benchmark problems in Section IV. Finally we conclude our work and propose some potential future work in Section V.

\section{Background}

SCP is a classical NP-hard combinatorial optimization problem, and has attracted the interests of researchers for several decades. Many exact algorithms, heuristic, and metaheuristic approaches have been proposed and reported in the literature.

Existing exact algorithms to solve SCP is mainly based on the branch-and-bound and branch-and-cut search algorithms \cite{BalachandarKannan2010MetaheuristicAlgorithm}. Fisher and Kedia proposed an exact branch-and-bound algorithm based on a dual heuristic \cite{FisherKedia1990OptimalSolutionSet}. This algorithm is capable of solving SCP instances with up to 200 elements and 2000 subsets (200$\times$2000). Beasley combined a Lagrangian-based heuristic, sub-gradient optimization, and linear programming to improve the branching strategy \cite{Beasley1987AlgorithmSetCovering}. He then enhanced this algorithm using feasible solution exclusion constraints and Gomory's f-cut in \cite{BeasleyJornsten1992EnhancingAlgorithmSet}. The algorithm is tested on instances with matrices up to the order of 400$\times$4000. Harche and Thompson developed a column subtraction exact algorithm to solve sparse instances of SCP up to the order of 900$\times$8000 \cite{HarcheThompson1994ColumnSubtractionAlgorithm:}. All the above exact algorithms are based on the tree-search algorithm, which has limitations such as extremely high computational complexity, very large searching space, and relatively poor performance \cite{BalachandarKannan2010MetaheuristicAlgorithm}. Due to these drawbacks, researchers resorted to approximate algorithms to meet the requirement of less stringent computation with satisfactory solution quality.

Greedy algorithms are one of the heuristic approaches to quickly solve large combinatorial problems. Chvatal proposed the very first greedy algorithm to solve SCP in 1979 \cite{Chvatal1979GreedyHeuristicSet}. However, due to its deterministic and myopic feature, this algorithm can rarely generate good solutions despite it is fast and simple. In order to improve the performance of the canonical greedy algorithm, researchers have tried to introduce some stochastic features into it \cite{FeoResende1989ProbabilisticHeuristicComputationally} \cite{HaouariChaouachi2002probabilisticgreedysearch}. Generally, greedy algorithms with stochastic features can generate better solutions than the canonical greedy ones. There are also some non-greedy-based heuristics. Caprara's work \cite{CapraraFischettiToth1999HeuristicMethodSet} is an example which gives good solutions. This work introduced variable fixing and pricing techniques into a Lagrangian heuristics.

The research on employing metaheuristic algorithms, especially evolutionary algorithms, to solve SCP has been intensely investigated in the last decade. A wide range of metaheuristics have been utilized, including genetic algorithm \cite{BeasleyChu1996GeneticAlgorithmSet} \cite{Aickelin2002IndirectGeneticAlgorithm}, ant colony optimization \cite{LessingDumitrescuStutzle2004ComparisonBetweenACO}, simulated annealing \cite{JacobsBrusco1995Note:Localsearch}, and artificial neural networks \cite{OhlssonPetersonSoderberg2001EfficientMeanField}. The high adaptability and superior performance of metaheuristic make it a competitive approach to solve SCP. Moreover, due to the characteristics of unicost SCP, researchers have proposed some algorithms to solve this special kind of SCP. Grossman and Wool designed an artificial neural network framework to solve unicost SCP \cite{GrossmanWool1997ComputationalExperienceWith}. However, the existing algorithms to solve SCP have two drawbacks in general. Firstly, most algorithms are designed to solve non-unicost SCP. Beasley and Chu pointed out that their algorithms based on Lagrangian relaxation and genetic algorithm were not recommended for unicost problems \cite{BeasleyChu1996GeneticAlgorithmSet}. From the literature, very few algorithms are found to work effectively for both unicost and non-unicost problem. Secondly, most algorithms which can generate satisfactory results are difficult to implement, while the simple algorithms, e.g., greedy algorithms, are less competitive in performance.

In order to overcome these two drawbacks, the goal of this work is to design a robust and simple metaheuristic based on CRO that can generate good results for both unicost and non-unicost SCP. CRO is a recently proposed general-purpose metaheuristic, which has been developed intensely in the past few years. CRO was proposed by Lam \textit{et al.} in 2010 \cite{LamLi2010ChemicalReactionInspired}, and was originally designed for solving combinatorial optimization problems. They solved some classical problems, e.g., quadratic assignment problem and channel assignment problem. CRO is also applied to solve some real world applications like real-time monitoring \cite{YuLiLam2012SensorDeploymentAir} and smart grid \cite{YuLiLam2013OptimalV2GScheduling}. For continuous problems, Lam \textit{et al.} also proposed a variant of CRO, i.e., real-coded CRO, which has demonstrated superior performance in many real-world applications, e.g., training artificial neural networks \cite{YuLamLi2011EvolutionaryArtificialNeural} and optimal power flow problem in power grid \cite{SunLamLiXuYu2012ChemicalReactionOptimization}.

\section{Algorithm Design}

In this section we will first introduce CRO. Then the implementation of operators used in CRO will be presented.

\subsection{Chemical Reaction Optimization}

We consider a number of molecules in a closed container with an attached energy buffer. Molecules are the basic operating agents of CRO. CRO manipulates and controls a collection of molecules to explore the solution space of an optimization problem. CRO considers the molecular structure as the feasible solution of the optimization problem. Besides the molecular structure, a molecule also has some other attributes that help the algorithm to perform optimization. Typically each molecule possesses two different kinds of energy, namely, potential energy (PE) and kinetic energy (KE). We use PE to represent the solution quality, or objective function value, of the corresponding molecular structure. The solution space is decribed by the potential energy surface (PES). The molecules can move freely on the PES. Every position on the PES is associated with a PE value. The lower the PE (for objective function minimization), the better is the solution. KE quantifies the ability of the molecule to move towards an area on the PES with higher values. The larger a molecule's KE, the higher it can position itself on the PES, which means that this molecule can accept worse solutions. This feature is very important for the cases when CRO tries to manipulate the molecules to jump out of a local optimum in the solution space.

CRO controls and manipulates the molecules with four different elementary reactions, namely, on-wall ineffective collision (\textit{on-wall}), decomposition (\textit{dec}), inter-molecular ineffective collision (\textit{inter}), and synthesis (\textit{syn}), each of which is described by an operator. Each operator modifies the molecular structures of some molecules, performing stochastic exploration or exploitation in the solution space. The four elementary reactions have different energy handling schemes and molecular structure operations. Meanwhile, they share a feature, which distinguishes CRO from other metaheuristics. All the operations in CRO must comply with the energy conservation law, which states that although energy is allowed to transform between types, the total energy in an isolated system (i.e. the container in CRO) shall remain constant. In CRO, the total energy of the system before and after an elementary reaction is constant. Interested reader can refer to \cite{LamLi2012ChemicalReactionOptimization:} for the detailed implementation of CRO.

\subsection{Encoding Scheme}

There are two major encoding schemes to solve SCP in the literature. The first one is the natural encoding scheme. This scheme uses a binary vector of length $n$ for a solution. Each element in the vector represents the status of one particular subset. In other words, this encoding scheme uses the $\mathsf{X}$ in (\ref{eqn:form2}) as the solution for optimization. This encoding scheme is easy to implement, but a random solution generated from this scheme is not guaranteed to be a cover of $\mathbb{M}$. In order to overcome this drawback, a second type of encoding scheme is developed for SCP.

The second encoding scheme, which we adopt to solve SCP with CRO, has each value in the solution vector representing a subset index. The values are selected from the indices of those subsets that cover the corresponding element. For example, consider a solution vector $[2,6,7,2,\cdots]$. This solution uses the second subset to cover the first and fourth elements in $\mathbb{M}$, and the sixth subset to cover the second element and so forth. Solutions in the second scheme have shorter length than those in the first in general (for most non-unicost SCP, $m$ is smaller than $n$). Moreover, all the solutions generated by this encoding scheme satisfy the constraints of SCP naturally \cite{BalachandarKannan2010MetaheuristicAlgorithm}.

\subsection{Algorithm Design}

CRO defines four different types of elementary reactions, which possess different functionalities. So we design a corresponding operator for each of them. We also design an initial solution generator to generate the solution structures of new molecules. We generate all random numbers uniformly in the solution space, unless stated otherwise.

\subsubsection{Initial Solution Generator}

This operator is applied whenever a new molecule is generated. Instead of randomly assigning subset indices (this is common when we solve other optimization problems), we use a reverse cumulative scheme to select a random subset to cover this element. The scheme is stated in Algorithm \ref{alg:init}.
\begin{algorithm}
	\caption{\sc{Reverse Cumulative Scheme}}
	\label{alg:init}
	\begin{algorithmic}[1]
		\ForAll{Elements $i$ in a solution}
			\State Find all subsets $\mathbb{X}_i$ that cover the i-th element in $\mathbb{M}$.
			\State Find the maximum cost $c_{max}$ and the minimum cost \par $c_{min}$ of subsets in $\mathbb{X}_i$.
			\ForAll{$S_j\subseteq\mathbb{X}_i$}
				\State Assign a reverse cumulative value $v_j$.
				\State Assign its probability of being selected $p_j$.
			\EndFor
			\State Select the value of Element $i$ according to each \par subsets' $p_j$.
		\EndFor
	\end{algorithmic}
\end{algorithm}

In Algorithm \ref{alg:init}, the variables are calculated by
\begin{equation}\label{eqn:rev1}
v_j=c_{max}+c_{min}-c_j
\end{equation}
and
\begin{equation}\label{eqn:rev2}
p_j=\frac{v_j}{\sum^{|\mathbb{X}_i|}_{k=1}v_k}
\end{equation}

This scheme renders the initial solution more likely to include those subsets with lower cost, while ensuring the solution complies with all constraints. This scheme assigns the subsets with lower cost a larger possibility of being selected by (\ref{eqn:rev1}) and (\ref{eqn:rev2}). For example, the first element in $\mathbb{M}$ can be covered by Subsets 1, 3, and 6, whose costs are 2, 4, and 5, respectively. Then the reverse cumulative value of Subset 1 is calculated by $5+2-2=5$, and that for Subsets 3 and 6 are 3 and 2. Thus the probability of being selected are 0.5, 0.3, and 0.2, respectively.

\subsubsection{Neighborhood Search Operator}

This operator is applied to the two ineffective elementary reactions, namely, on-wall and inter. This operator modifies the input solution slightly to perform a local search. We employ a perturbation heuristic to act as the neighborhood search operator in CRO. The algorithm is stated in Algorithm \ref{alg:neigh}.
\begin{algorithm}
	\caption{\sc{Perturbation Heuristic}}
	\label{alg:neigh}
	\begin{algorithmic}[1]
		\ForAll{Subsets $S_j$ in the solution}
			\State Calculate its cost efficiency value $e^c_j$.
		\EndFor
		\State Find the Subset $S_i$ with the lowest cost efficiency.
		\State Remove all $i$ in the solution and leave blanks.
		\While{There are blanks in the solution}
			\ForAll{Subsets $S_k\subseteq\mathbb{N}$}
				\State Calculate its repair efficiency value $e^r_k$
			\EndFor
			\ForAll{Subsets $S_k\subseteq\mathbb{N}$}
				\State Calculate its probability of being selected $p_k$
			\EndFor
			\State Select a subset $S_l$to repair the solution according to \par each subsets' $p_k$.
			\State Use $l$ to fill all the blanks which can be covered by \par $S_l$.
		\EndWhile
	\end{algorithmic}
\end{algorithm}

In Algorithm \ref{alg:neigh}, the variables are calculated by
\begin{equation}\label{eqn:heu1}
e^c_j=\frac{n_j}{c_j},
\end{equation}
where $n_j$ is the occurrence of $j$ in the solution, and
\begin{equation}\label{eqn:heu2}
e^r_k=\frac{s_k}{c_k},
\end{equation}
where $s_k$ is the number of blanks that Subset $S_k$ can cover in the solution, and
\begin{equation}\label{eqn:heu3}
p_k=\frac{e^r_k}{\sum^{|\mathbb{X}|}_{j=1}e^r_j}
\end{equation}

This perturbation heuristic can be divided into two major parts: remove and repair. In the remove phase (Lines 1 to 5 of Algorithm \ref{alg:neigh}), the least efficient subset is removed from the solution. Its cost efficiency value is calculated by (\ref{eqn:heu1}). Assume a Subset 2 with cost 10. If it covers two elements in the solution, its cost efficiency value is 5. At the end of the remove phase, we have an incomplete solution with one or several blank positions. Then in the repair phase, we select a most efficient subset to fill in the blank(s). This repair efficiency value is calculated by (\ref{eqn:heu2}). With the repair efficiency values, we can calculate the probability of being selected for repairing using (\ref{eqn:heu3}). For example, assume we have an incomplete solution $[1,\_,4,\_,\_,\cdots]$ where the underline positions are blank positions. Subset 3 can cover two blanks with a cost 20, and Subset 5 can cover three blanks with a cost 40. So the repair efficiency values of Subsets 3 and 5 are 0.1 and 0.075, respectively. If there are no other subsets, the probabilities of selecting Subsets 3 and 5 are 57\% and 43\%, respectively. If the algorithm chooses Subset 3 for repairing, the two blanks which can be covered by Subset 3 is filled with 3. This completes one iteration of the repairing phase and this process iterates until all blanks are filled.

For on-wall,  we directly employ this perturbation scheme to the molecule, while for inter, we apply the scheme to the two input molecules simultaneously.

\begin{table}[t]
  \centering
  \caption{Beasley's OR Library Non-unicost Instances}
    \begin{tabular}{rrrrrr}
    \hline
    Set & Instances & Size ($m\times n$) & Costs & Density & Opt. Solution \\
    \hline
    4 & 10 & 200$\times$1000 & 1-100 & 2\% & Known \\
    5 & 10 & 200$\times$2000 & 1-100 & 2\% & Known \\
    6 & 5 & 200$\times$1000 & 1-100 & 5\% & Known \\
    A & 5 & 300$\times$3000 & 1-100 & 2\% & Known \\
    B & 5 & 300$\times$3000 & 1-100 & 5\% & Known \\
    C & 5 & 400$\times$4000 & 1-100 & 2\% & Known \\
    D & 5 & 400$\times$4000 & 1-100 & 5\% & Known \\
    NRE & 5 & 500$\times$5000 & 1-100 & 10\% & Unknown \\
    NRF & 5 & 500$\times$5000 & 1-100 & 20\% & Unknown \\
    NRG & 5 & 1000$\times$10000 & 1-100 & 2\% & Unknown \\
    NRH & 5 & 1000$\times$10000 & 1-100 & 5\% & Unknown \\
    \hline
    \end{tabular}
  \label{tab:orlib1}
\end{table}

\subsubsection{Dec and Syn Operators}

The main purpose of dec and syn is to help the molecules jump out of local optimum. So we usually impose severe changes to the input molecule(s). For dec operator, we first copy the input molecule to the two output molecules, then perform the neighborhood search operator on each molecule for 10 times separately. The resultant molecules are regarded as the final output molecules. For the syn operator, we introduce a probabilistic combination scheme to combine the two input molecules and create a new one. Assume the two input solutions are $\mathsf{X}_1$ and $\mathsf{X}_2$, and their costs are $c_1$ and $c_2$, respectively. Each element in the output new solution is drawn from the same position of either $\mathsf{X}_1$ or $\mathsf{X}_2$ with the probability of $\frac{c_2}{c_1+c_2}$ and $\frac{c_1}{c_1+c_2}$, respectively.

\section{Simulation Results}

In this section we will first introduce the benchmark instances used to evaluate the performance of our proposed hCRO algorithm. Then the detailed simulation parameter settings, results, and comparisons are presented.

\subsection{Benchmark Instances}

\begin{table}[t]
  \centering
  \caption{Beasley's OR Library Unicost Instances}
    \begin{tabular}{rrrrrr}
    \hline
    Set & Instances & Size ($m\times n$) & Density & Opt. Solution \\
    \hline
    E & 5 & 50$\times$500 & 2\% & Known \\
    CLR.10 & 1 & 511$\times$210 & 2\% & Unknown \\
    CLR.11 & 1 & 1023$\times$330 & 5\% & Unknown \\
    CLR.12 & 1 & 2047$\times$495 & 2\% & Unknown \\
    CLR.13 & 1 & 4095$\times$715 & 5\% & Unknown \\
    CYC.6 & 1 & 240$\times$192 & 2\% & Known \\
    CYC.7 & 1 & 672$\times$448 & 5\% & Unknown \\
    CYC.8 & 1 & 1792$\times$1024 & 10\% & Unknown \\
    CYC.9 & 1 & 4608$\times$2304 & 20\% & Unknown \\
    CYC.10 & 1 & 11520$\times$5120 & 2\% & Unknown \\
    CYC.11 & 1 & 28160$\times$11264 & 5\% & Unknown \\
    \hline
    \end{tabular}
  \label{tab:orlib2}
\end{table}

\begin{table}[t]
  \centering
  \caption{hCRO Parameter Values}
    \begin{tabular}{rr}
    \hline
    Parameter & Value \\
    \hline
    Initial population size & 10 \\
    Initial molecular kinetic energy & 1000 \\
    Initial central energy buffer size & 10000 \\
    Collision rate & 0.1 \\
    Energy loss rate & 0.1 \\
    Decomposition threshold & 10000 \\
    Synthesis threshold & 1000 \\
    \hline
    \end{tabular}
  \label{tab:param}
\end{table}

We will test the performance of our proposed hCRO using 65 non-unicost SCP test instances from Beasley's OR Library \cite{Beasley1990LagrangianHeuristicSet}. We note that almost all SCP algorithms developed in the past two decades were tested using these problems. The instances are divided into 11 different sets, as listed in Table \ref{tab:orlib1}, where the density is the percentage of non-zero entries in the SCP matrix $\mathbf{A}$.

We also test the performance of hCRO with the unicost instances in Beasley's OR Library and the information about these instances are listed in Table \ref{tab:orlib2}.

\subsection{Parameter Tuning and Simulation Environment}

When applying hCRO to perform simulation, several parameters must be set. Coy \textit{et al.} note that it is often very difficult to find appropriate parameter settings for metaheuristics, and common procedures of generating proper parameter values have ranged from simple trial-and-error to complicated sensitivity analysis \cite{CoyGoldenRungerWasil2001UsingExperimentalDesign}. In this work, we use the trial-and-error method to tune the hCRO parameters, as in some previous CRO efforts \cite{YuLiLam2012SensorDeploymentAir} \cite{YuLamLi2012RealcodedChemical}. The first instances in the SCP problem set 4, 5, 6, A, B, C, and D are selected as representative instance for parameter tuning and we perform a series of test runs on these instances, using the methodology described in \cite{YuLamLi2012RealcodedChemical}. The final parameters used for all test instances are listed in Table \ref{tab:param}.

Our proposed approach was implemented in C++ on a computer with an Intel Core i5 3.1-GHz processor and MinGW compiler. In our experimental study, 100 trials of hCRO were conducted for each of these test problems. The maximum function evaluation limit is set to $n\times1000$, which is smaller or equal to that in all the metaheuristics we compare with.

\subsection{Simulation Results and Comparison with Other Algorithms}

\begin{table}
    \centering
    \caption{hCRO Results on Non-unicost Instances}
    \begin{tabular}{l|r|r|rr|rr|rr}
        \hline
        Inst. & BKS & Opt. & \multicolumn{2}{c|}{Best} & \multicolumn{2}{c|}{Mean} & \multicolumn{2}{c}{Worst} \\ \cline{4-9}
        & & & Value & Pct. & Value & Pct. & Value & Pct. \\\hline
        4.1   & 429 & 100 & 429 & 0 & 429 & 0 & 429 & 0 \\ 
        4.2   & 512 & 100 & 512 & 0 & 512 & 0 & 512 & 0 \\ 
        4.3   & 516 & 100 & 516 & 0 & 516 & 0 & 516 & 0 \\ 
        4.4   & 494 & 100 & 494 & 0 & 494 & 0 & 494 & 0 \\ 
        4.5   & 512 & 100 & 512 & 0 & 512 & 0 & 512 & 0 \\ 
        4.6   & 560 & 100 & 560 & 0 & 560 & 0 & 560 & 0 \\ 
        4.7   & 430 & 100 & 430 & 0 & 430 & 0 & 430 & 0 \\ 
        4.8   & 492 & 100 & 492 & 0 & 492 & 0 & 492 & 0 \\ 
        4.9   & 641 & 100 & 641 & 0 & 641 & 0 & 641 & 0 \\ 
        4.10  & 514 & 100 & 514 & 0 & 514 & 0 & 514 & 0 \\ \hline
        5.1   & 253 & 100 & 253 & 0 & 253 & 0 & 253 & 0 \\ 
        5.2   & 302 & 100 & 302 & 0 & 302 & 0 & 302 & 0 \\ 
        5.3   & 226 & 100 & 226 & 0 & 226 & 0 & 226 & 0 \\ 
        5.4   & 242 & 100 & 242 & 0 & 242 & 0 & 242 & 0 \\ 
        5.5   & 211 & 100 & 211 & 0 & 211 & 0 & 211 & 0 \\ 
        5.6   & 213 & 100 & 213 & 0 & 213 & 0 & 213 & 0 \\ 
        5.7   & 293 & 100 & 293 & 0 & 293 & 0 & 293 & 0 \\ 
        5.8   & 288 & 100 & 288 & 0 & 288 & 0 & 288 & 0 \\ 
        5.9   & 279 & 100 & 279 & 0 & 279 & 0 & 279 & 0 \\ 
        5.10  & 265 & 100 & 265 & 0 & 265 & 0 & 265 & 0 \\ \hline
        6.1   & 138 & 100 & 138 & 0 & 138 & 0 & 138 & 0 \\ 
        6.2   & 146 & 100 & 146 & 0 & 146 & 0 & 146 & 0 \\ 
        6.3   & 145 & 100 & 145 & 0 & 145 & 0 & 145 & 0 \\ 
        6.4   & 131 & 100 & 131 & 0 & 131 & 0 & 131 & 0 \\ 
        6.5   & 161 & 100 & 161 & 0 & 161 & 0 & 161 & 0 \\ \hline
        A.1   & 253 & 100 & 253 & 0 & 253 & 0 & 253 & 0 \\ 
        A.2   & 252 & 100 & 252 & 0 & 252 & 0 & 252 & 0 \\ 
        A.3   & 232 & 100 & 232 & 0 & 232 & 0 & 232 & 0 \\ 
        A.4   & 234 & 100 & 234 & 0 & 234 & 0 & 234 & 0 \\ 
        A.5   & 236 & 100 & 236 & 0 & 236 & 0 & 236 & 0 \\ \hline
        B.1   & 69  & 100 & 69  & 0 & 69  & 0 & 69  & 0 \\ 
        B.2   & 76  & 100 & 76  & 0 & 76  & 0 & 76  & 0 \\ 
        B.3   & 80  & 100 & 80  & 0 & 80  & 0 & 80  & 0 \\ 
        B.4   & 79  & 100 & 79  & 0 & 79  & 0 & 79  & 0 \\ 
        B.5   & 72  & 100 & 72  & 0 & 72  & 0 & 72  & 0 \\ \hline
        C.1   & 227 & 100 & 227 & 0 & 227 & 0 & 227 & 0 \\ 
        C.2   & 219 & 100 & 219 & 0 & 219 & 0 & 219 & 0 \\ 
        C.3   & 243 & 100 & 243 & 0 & 243 & 0 & 243 & 0 \\ 
        C.4   & 219 & 100 & 219 & 0 & 219 & 0 & 219 & 0 \\ 
        C.5   & 215 & 100 & 215 & 0 & 215 & 0 & 215 & 0 \\ \hline
        D.1   & 60  & 100 & 60  & 0 & 60  & 0 & 60  & 0 \\ 
        D.2   & 66  & 100 & 66  & 0 & 66  & 0 & 66  & 0 \\ 
        D.3   & 72  & 100 & 72  & 0 & 72  & 0 & 72  & 0 \\ 
        D.4   & 62  & 100 & 62  & 0 & 62  & 0 & 62  & 0 \\ 
        D.5   & 61  & 100 & 61  & 0 & 61  & 0 & 61  & 0 \\ \hline
        NRE.1 & 29  & 100 & 29  & 0 & 29  & 0 & 29  & 0 \\ 
        NRE.2 & 30  & 100 & 30  & 0 & 30  & 0 & 30  & 0 \\ 
        NRE.3 & 27  & 100 & 27  & 0 & 27  & 0 & 27  & 0 \\ 
        NRE.4 & 28  & 100 & 28  & 0 & 28  & 0 & 28  & 0 \\ 
        NRE.5 & 28  & 100 & 28  & 0 & 28  & 0 & 28  & 0 \\ \hline
        NRF.1 & 14  & 100 & 14  & 0 & 14  & 0 & 14  & 0 \\
        NRF.2 & 15  & 100 & 15  & 0 & 15  & 0 & 15  & 0 \\ 
        NRF.3 & 14  & 100 & 14  & 0 & 14  & 0 & 14  & 0 \\ 
        NRF.4 & 14  & 100 & 14  & 0 & 14  & 0 & 14  & 0 \\ 
        NRF.5 & 13  & 100 & 13  & 0 & 13  & 0 & 13  & 0 \\  \hline
        NRG.1 & 176 & 100 & 176 & 0 & 176 & 0 & 176 & 0 \\ 
        NRG.2 & 154 & 100 & 154 & 0 & 154 & 0 & 154 & 0 \\ 
        NRG.3 & 166 & 100 & 166 & 0 & 166 & 0 & 166 & 0 \\ 
        NRG.4 & 168 & 100 & 168 & 0 & 168 & 0 & 168 & 0 \\ 
        NRG.5 & 168 & 100 & 168 & 0 & 168 & 0 & 168 & 0 \\  \hline
        NRH.1 & 63  & 100 & 63  & 0 & 63  & 0 & 63  & 0 \\ 
        NRH.2 & 63  & 100 & 63  & 0 & 63  & 0 & 63  & 0 \\ 
        NRH.3 & 59  & 100 & 59  & 0 & 59  & 0 & 59  & 0 \\ 
        NRH.4 & 58  & 100 & 58  & 0 & 58  & 0 & 58  & 0 \\ 
        NRH.5 & 55  & 100 & 55  & 0 & 55  & 0 & 55  & 0 \\ 
        \hline
    \end{tabular}
  \label{tab:cro1}
\end{table}

\begin{table*}
    \centering
    \caption{hCRO Results on Non-unicost Instances}
    \begin{tabular}{l|r|r|rr|rr|rr}
        \hline
        Inst. & BKS & Opt. & \multicolumn{2}{c|}{Best} & \multicolumn{2}{c|}{Mean} & \multicolumn{2}{c}{Worst} \\ \cline{4-9}
        & & & Value & Pct. & Value & Pct. & Value & Pct. \\\hline
        E.1    & 5    & 100 & 5    & 0      & 5      & 0      & 5    & 0      \\
        E.2    & 5    & 100 & 5    & 0      & 5      & 0      & 5    & 0      \\ 
        E.3    & 5    & 100 & 5    & 0      & 5      & 0      & 5    & 0      \\ 
        E.4    & 5    & 100 & 5    & 0      & 5      & 0      & 5    & 0      \\ 
        E.5    & 5    & 100 & 5    & 0      & 5      & 0      & 5    & 0      \\ \hline
        CLR.10 & 25   & 100 & 25   & 0      & 25     & 0      & 25   & 0      \\ 
        CLR.11 & 23   & 100 & 23   & 0      & 23     & 0      & 23   & 0      \\ 
        CLR.12 & 23   & 100 & 23   & 0      & 23     & 0      & 23   & 0      \\ 
        CLR.13 & 23   & 100 & 23   & 0      & 23     & 0      & 23   & 0      \\ \hline
        CYC.6  & 60   & 100 & 60   & 0      & 60     & 0      & 60   & 0      \\ 
        CYC.7  & 144  & 11  & 144  & 0      & 146.7  & 0.0188 & 152  & 0.0556 \\ 
        CYC.8  & 344  & 14  & 344  & 0      & 349.2  & 0.0151 & 361  & 0.0494 \\ 
        CYC.9  & 780  & 0   & 789  & 0.0115 & 797.3  & 0.0221 & 819  & 0.05   \\ 
        CYC.10 & 1792 & 0   & 1802 & 0.0056 & 1832.6 & 0.0226 & 1872 & 0.0446 \\ 
        CYC.11 & 4103 & 0   & 4113 & 0.0024 & 4159.2 & 0.0137 & 4201 & 0.0239 \\ 
        \hline
    \end{tabular}
  \label{tab:cro2}
\end{table*}

The simulation results are presented in Table \ref{tab:cro1}. In this table, ``Inst.'' is the instance index,  ``BKS" is the optimum solution or the best known value, ``Opt." is the number of trials that hCRO is able to find the optimum solution or the best known value. We also present the mean, best, and worst objective function values of the 100 trials, and their respective percentages above the optimal value. The results of hCRO on unicost SCP instances are also presented in Table \ref{tab:cro2}.

\begin{table*}
    \centering
    \caption{Performance Gap Comparison on Non-unicost Instances (\%)}
    \begin{tabular}{l|rrrrrrrrr}
        \hline
        Problem Set & hCRO   & RaPS  & GRA   & CFT   & BeCh  & IGA   & PROG  & Be    & Greedy \\ \hline
        4           & 0.00  & 0.00  & 0.00  & 0.00  & 0.00  & 0.00  & 0.57  & 0.06  & 3.78   \\ 
        5           & 0.00  & 0.00  & 0.00  & 0.00  & 0.09  & 0.00  & 0.88  & 0.18  & 5.51   \\ 
        6           & 0.00  & 0.00  & 0.00  & 0.00  & 0.00  & 0.00  & 0.69  & 0.56  & 7.22   \\ 
        A           & 0.00  & 0.00  & 0.00  & 0.00  & 0.00  & 0.00  & 0.75  & 0.82  & 5.61   \\ 
        B           & 0.00  & 0.00  & 0.00  & 0.00  & 0.00  & 0.00  & 0.00  & 0.81  & 5.57   \\ 
        C           & 0.00  & 0.00  & 0.00  & 0.00  & 0.00  & 0.00  & 0.87  & 1.93  & 6.88   \\ 
        D           & 0.00  & 0.00  & 0.00  & 0.00  & 0.00  & 0.32  & 0.00  & 2.75  & 9.79   \\ 
        NRE         & 0.00  & 0.00  & 0.00  & 0.00  & 0.00  & 0.00  & 0.00  & 3.50  & 12.75  \\ 
        NRF         & 0.00  & 0.00  & 0.00  & 0.00  & 0.00  & 0.00  & 1.43  & 7.16  & 12.98  \\ 
        NRG         & 0.00  & 0.00  & 0.00  & 0.00  & 0.13  & 0.13  & 1.18  & 4.83  & 8.49   \\ 
        NRH         & 0.00  & 0.00  & 0.00  & 0.00  & 0.63  & 1.30  & 1.68  & 8.12  & 11.78  \\ \hline
        Overall     & 0.00  & 0.00  & 0.00  & 0.00  & 0.08  & 0.16  & 0.72  & 2.36  & 8.21   \\ \hline
        Opt. Found  & 65/65 & 65/65 & 65/65 & 65/65 & 61/65 & 61/65 & 22/65 & 20/65 & 0/65   \\
        \hline
    \end{tabular}
  \label{tab:comp1}
\end{table*}

From Tables \ref{tab:cro1} and \ref{tab:cro2}, we can see that hCRO has excellent performance in non-unicost instances, and all 65 optimum solutions are generated in every run for all instances. For unicost-SCP instances, hCRO obtains 12 optimums out of 15. For the remaining instances, hCRO can also generate a satisfactory result (error percentage around 1\% of best results).

In order to further demonstrate the performance of hCRO, we also compare hCRO with other algorithms proposed in the recent literature. We compare the performance on non-unicost instances of hCRO with the Lagrangian heuristic by Beasley (Be) \cite{Beasley1990LagrangianHeuristicSet}, the genetic algorithm by Beasley and Chu (BeCh) \cite{BeasleyChu1996GeneticAlgorithmSet}, the Lagrangian heuristic by Caprara \textit{et al.} (CFT) \cite{CapraraFischettiToth1999HeuristicMethodSet}, a probabilistic greedy search heuristic by Haouari and Chaouachi (PROG) \cite{HaouariChaouachi2002probabilisticgreedysearch}, an indirect genetic algorithm by Aickelin (IGA) \cite{Aickelin2002IndirectGeneticAlgorithm}, a metaheuristic for randomized priority search by Lan \textit{et al.} (RaPS) \cite{LanDePuyWhitehouse2007EffectiveAndSimple}, and a metaheuristic algorithm based on gravity by Balachandar and Kannan (GRA) \cite{BalachandarKannan2010MetaheuristicAlgorithm}. The comparison is presented in Table \ref{tab:comp1}, where the row of ``Opt. Found" is the number of instances in which the corresponding algorithm finds the optimal value or the best known solution out of the all 65 instances. The table shows the average gap of the global optimal found by the algorithm and BKS. For example, an algorithm generates an optimal value of 52 on a problem instance with a best known value of 50, the gap is defined as $(52-50)/50=4\%$. From the table we can see, hCRO is one of the best four algorithms that can find the optimal value or best known solution 100\% of the time.

We further demonstrate the performance of hCRO by comparing the simulation results on unicost instances with other algorithms. These algorithms include the heuristic random approximation by Peleg \textit{et al.} (RR) \cite{PelegSchechtmanWool1997RandomizedApproximationBounded}, a greedy heuristic by Chvatal (Gr) \cite{Chvatal1979GreedyHeuristicSet}, three algorithms proposed by Grossman and Wool in \cite{GrossmanWool1997ComputationalExperienceWith} (Alt-Gr, NN, and R-Gr), and a mean-field approach by Ohlsson \textit{et al.} (MF) \cite{OhlssonPetersonSoderberg2001EfficientMeanField}. The results are presented in Table \ref{tab:comp2}. From the comparison we can see hCRO again outperforms all other algorithms dramatically. Therefore, hCRO is an effective algorithm in solving both non-unicost and unicost set covering problems.

\begin{table*}
    \centering
    \caption{Performance Gap Comparison on Unicost Instances (\%)}
    \begin{tabular}{l|rrrrrrrrr}
        \hline
        Problem Set & hCRO   & RR  & Gr   & Alt-Gr  & NN  & R-Gr   & MF  \\ \hline

        Overall & 0.62   & 20.34  & 10.50  & 7.29  & 10.07  & 9.85  & 6.51    \\ \hline
    \end{tabular}
  \label{tab:comp2}
\end{table*}

\subsection{Analysis on Performance Contribution of the Proposed Heuristic Schemes and CRO Framework}

In order to analyze the contribution of different proposed heuristic schemes and the CRO framework to the outstanding performance of hCRO, we also construct different hCRO variations and a heuristic-based Genetic Algorithm (hGA).

To determine the impact of our initialization operator (Algorithm \ref{alg:init}) and our neighborhood search operator (Algorithm \ref{alg:neigh}), we create alternative operators for these two schemes and construct two new hCRO variations:

\begin{itemize}
\item hCRO/IR: This variation of hCRO utilizes our previously proposed perturbation heuristic as the neighborhood search operator, with a ``random pick scheme'' as the initial population generator. In the ``random pick scheme'', we first construct a permutation vector of all elements in $\mathbb{M}$. We then start from the first element $i$ in this permutation vector and find all subsets $\mathbb{X}_i$ that cover $i$. Then we randomly select one such subset and go on to the next element $j$. If $j$ is not covered yet, we repeat the previous step, i.e. find all subsets that cover $j$ and randomly select one. If $j$ is already covered, we go to the next element. This process repeats until all elements are covered by at least one subset.
\item hCRO/NR: This variation of hCRO utilizes our previously proposed reverse cumulative scheme as the initial population generator, with a ``remove-repair scheme'' as the neighborhood search operator. In the ``remove-repair scheme'', we first randomly remove one subset in the solution. Then we determine all elements not yet covered by any subsets. The remaining part is similar to the ``random pick scheme'', where we build a permutation vector, find all subsets that cover the current element and randomly select one such subset.
\end{itemize}

Besides hCRO/IR and hCRO/NR, we also develop hGA to solve SCP in order to reveal the contribution of the CRO framework in hCRO:
\begin{itemize}
\item hGA: We utilize our previously proposed reverse cumulative scheme as the initial population generator. We use the inter-molecular ineffective collision operator in hCRO as the crossover operator in hGA, i.e., two perturbation heuristics are executed simultaneously. As for the mutation operator, we adopt the decomposition operator in hCRO and select the better-performing solution as the mutated chromosome. We set the population size the same with hCRO. The crossover rate and mutation rate are set at 0.8 and 0.2, respectively, which is a commonly used combination of parameters for GA.
\end{itemize}

We perform simulations of hCRO, hCRO/IR, hCRO/NR, and hGA on both unicost and non-unicost instances. The simulation results are presented in Table \ref{tab:comp3}.

\begin{table}
    \centering
    \caption{Performance Gap Comparison on hCRO, hCRO/IR, hCRO/NR, and hGA (\%)}
    \begin{tabular}{l|rrrr}
        \hline
        Problem Set & hCRO   & hCRO/IR & hCRO/NR & hGA  \\ \hline
        Non-unicost & 0.00   & 0.04  & 1.78  & 0.11 \\
        Unicost & 0.62   & 2.19  & 14.64  & 3.82 \\ \hline
    \end{tabular}
  \label{tab:comp3}
\end{table}

From the results we can see that hCRO always performs the best. While hCRO/IR can generate similar results as hCRO, there is still a small gap between the performance of the two algorithms. This demonstrates the superiority of the reverse cumulative scheme. hCRO/NR performs much worse than all other algorithms, and this observation underlines the importance of the perturbation heuristic in generating good results. The results also show that the CRO framework is superior to conventional problem solver frameworks like GA. This is probably due to the unique energy conservation design in CRO as well as its tolerant, energy-related individual selection pattern \cite{LamLi2010ChemicalReactionInspired}.

\section{Conclusion}

In this paper we develop a heuristic-based CRO algorithm to solve non-unicost and unicost SCP. This algorithm introduces two heuristics into the operators of CRO. We study the performance of hCRO with a series of benchmark test instances from the Beasley's OR Library and show that hCRO enjoys superior performance in terms of the solution quality when compared with other algorithms. hCRO is able to find all 65 optimal solutions in non-unicost instances and it demonstrates outstanding performance when applied to unicost SCP. We also perform a series of test over different variations of hCRO as well as a heuristic-based GA to demonstrate the contribution of these two heuristics and the CRO framework on the final performance.

Further research includes the improvement of the performance for huge SCP problems. We can also introduce some implementation techniques that are commonly used by other SCP algorithms, e.g., group memory between iterations, to CRO. Some of the heuristics developed in our proposed algorithm can also be applied to solve other applications of CRO, such as the bin packing problem and the multi-dimensional knapsack problem. Last but not least, we can use the proposed algorithm to solve real-world applications of SCP.

\section*{Acknowledgement}
This research is supported in part by the University of Hong Kong Strategic Research Theme on Computation and Information. A.Y.S. Lam is supported in part by the Faculty Research Grant of Hong Kong Baptist University, under Grant No. FRG2/13-14/045.

\bibliographystyle{IEEEtran}
\bibliography{IEEEabrv,../../../bib/publications}

\end{document}